# Advancements in Machine Learning and Deep Learning for Early Detection and Management of Mental Health Disorder

Kamala Devi Kannan[a], Senthil Kumar Jagatheesaperumal[b,*], Rajesh N. V. P. S. Kandala[c], Mojtaba Lotfaliany[d], Roohallah Alizadehsani[d,e] and Mohammadreza Mohebbi[f]

[a]Department of Computer Science and Engineering, Mepco Schlenk Engineering College, Sivakasi, 626005, Tamil Nadu, India
[b]Department of Electronics and Communication Engineering, Mepco Schlenk Engineering College, Sivakasi, 626005, Tamil Nadu, India
[c]School of Electronics Engineering (SENSE), VIT-AP University, Amaravati, Andhra Pradesh, India
[d]The Institute for Mental and Physical Health and Clinical Translation (IMPACT), School of Medicine, Deakin University, Geelong, Australia
[e]Biostatistics Unit, Faculty of Health, Deakin University, Geelong, Australia
[f]School of Medicine, Deakin University, Geelong, Australia

## Abstract

For the early identification, diagnosis, and treatment of mental health illnesses, the integration of deep learning (DL) and machine learning (ML) have started playing a significant role. By evaluating complex data from imaging, genetics, and behavioral assessments, these technologies have the potential to improve clinical results significantly. However, they also present unique challenges relating to data integration and ethical issues. The development of ML and DL methods for the early diagnosis and treatment of mental health issues is reviewed in this survey. It examines a range of applications, with a particular emphasis on behavioral assessments, genetic and biomarker analysis, and medical imaging for the diagnosis of diseases like depression, bipolar disorder, and schizophrenia. Predictive modeling for illness development is further discussed in the review, focusing on the function of risk prediction models and longitudinal investigations. Important discoveries show how ML and DL might improve treatment outcomes and diagnostic accuracy while tackling methodological inconsistency, data integration, and ethical concerns. The study emphasizes the significance of building real-time monitoring systems for individualized treatment, improving data fusion techniques, and interdisciplinary collaboration. Upcoming studies should concentrate on surmounting these obstacles to maximize ML and DL's valuable and moral implementation in mental health services.

## 1. Introduction

Mental health disorders remain a significant global challenge, affecting an individual's reasoning, emotions, and social behavior [1]. Traditional diagnostic methods often fail to capture the complex underlying causes of mental illnesses, leading to misinterpretation and inadequate interventions. Mental health disorders can have severe consequences, and novel techniques are required for effective intervention and prevention. These strategies can only be achieved through early diagnosis, significantly improving treatment outcomes. Machine Learning (ML) and Deep Learning (DL) techniques offer a promising path for early diagnosis by analyzing complex patterns in patient data and providing actionable insights [2].

Psychiatric conditions can result in severe impairments, but with the right treatment and care, many patients can recover from emotional or mental disorders. However, traditional diagnostic techniques rely heavily on self-reports and questionnaires to identify social and emotional patterns [3]. ML techniques can overcome these limitations by analyzing complex data such as brain scans, genetic markers, and behavioral assessments [4]. ML methods, particularly in supervised learning, have demonstrated success in diagnosing mental disorders through their ability to handle large datasets, learning patterns, and performing classification tasks [5]. It is challenging to discriminate between bipolar and unipolar depression on a clinical level alone from neurobiological markers such as those collected through functional and structural MRI. The neuroimaging studies found in [6], differentiates in functional and structural brain abnormalities in patients with unipolar versus bipolar depression particularly affecting emotion- and reward-processing networks such as the amygdala and prefrontal cortex.

*Corresponding author
**Principal corresponding author
kamaladevik@mepcoeng.ac.in (K.D. Kannan); senthilkumarj@mepcoeng.ac.in (S.K. Jagatheesaperumal); rajesh.kandala@vitap.ac.in (R.N.V.P.S. Kandala); mojtaba.lotfaliany@deakin.edu.au (M. Lotfaliany); r.alizadehsani@deakin.edu.au (R. Alizadehsani); m.mohebbi@deakin.edu.au (M. Mohebbi)
ORCID(s):

ML has been widely applied in three categories: clustering, classification, and regression [3]. These techniques mimic human learning by analyzing data to gradually refine the accuracy of predictions. ML is particularly effective for understanding brain patterns, genetic predispositions, and behavioral indicators, offering a valuable tool for psychiatrists and researchers. Scientific evidence indicates that psychiatric disorders can be better predicted using ML than with traditional diagnostic frameworks such as ICD or DSM [4].

Supervised learning is a major ML technique used for mental health diagnosis. It works by mapping input variables to output variables, making predictions on unseen data. One common supervised learning model is the Support Vector Machine (SVM), which excels at both structured and semi-structured data classification tasks [7]. While SVM is advantageous in terms of generalization and minimizing overfitting, its performance can degrade when handling noisy or large datasets, as training time increases with dataset size. Another supervised method, decision trees, uses decision rules based on target variable features to predict outcomes [8]. Logistic regression is also widely used in predicting binary outcomes, and Naive Bayes employs probabilistic approaches for classification based on the assumption of feature independence [9]. New developments like federated deep learning to real-world deployment suggest the continuous evolution of ML in a fast-paced research environment. To illustrate this, a framework of multi-site federated domain adaptation via Transformer (FedDAvT) was presented in [10] to perform Alzheimer's classification by addressing site-specific data heterogeneity and protecting patient privacy. It was shown that by aligning self-attention maps between domains, strong classification accuracy can be achieved over distributed data without sharing raw patient information. Ensemble learning is another widely applied ML strategy, combining several learners into a single model to improve predictive accuracy. Techniques like stacking, bagging, and boosting allow multiple models to work together to enhance performance [11]. For example, Random Forest (RF) is a bagging technique that generates multiple decision trees to create a final model by aggregating their results. Similarly, Extreme Gradient Boosting (XGBoost) uses gradient boosting to achieve scalable, distributed learning for classification tasks. Transfer learning, which transfers knowledge from one task to improve performance on another related task, is also gaining attention in mental health research [12]. Although these techniques offer promising results, they raise ethical and legal concerns regarding patient data anonymization and privacy.

The generalizability of a model across different clinical sites, and how AI can be implemented into practice, also continue to be of key interest. In study [13] the authors used longitudinal EHR data to predict transition to diagnoses of schizophrenia or bipolar disorder. While they were able to predict schizophrenia specifically, the overall XGBoost model dropped from an AUROC of 0.70 on training data to 0.64 on hospital sites, indicating challenges with spatial generalizability across institutions. The authors specifically noted clinical notes to be valuable for this predictive task. To overcome challenges with clinical adoption, frameworks like an interactive, interpretable AI copilot have been developed. AICare [14] introduced an AI copilot that grounds EHR-based risk predictions in explainable visualizations and LLM-generated diagnosis recommendations. Unlike other work that focuses on expedience in triage, their clinical study indicated an improved cognitive workload, and showed that trust is context-dependent, built upon verification. Mitigation strategies include online learning algorithms like elastic weight consolidation [15].

The affective disorders (MDD, BD) are clinically heterogeneous with overlapping symptoms, trajectories, and comorbidity patterns, and thus the differential diagnosis is challenging and requires personalized treatment choices. Recent works combining neuroimaging, neuroendocrine, inflammatory, and cognitive biomarkers reinforce the heterogeneity, the complexity, and the inefficacy of symptom-based classification alone, in MDD and BD [16, 17]. These works emphasize that mood disorders are a result of multiple pathways dysregulation, therefore high-dimensional models such as ML and DL, that have the ability to grasp complex high-dimensional patterns and improve classification/risk evaluation are appropriate.

### 1.1. Related reviews

Researchers in this field have now extended numerous ML techniques to DL. In a review by [18], the authors exploited predictions on mental health issues using ML techniques and discussed future research directions. Thirty articles were collected for systematic review using the PRISMA methodology. The research articles were categorized based on different mental health disorders, such as post-traumatic stress disorder, depression, anxiety, bipolar disorder, and schizophrenia among children.

In another study [19], the authors examined 33 articles on the detection of attention deficit hyperactivity dis-order (ADHD), anorexia nervosa, post-traumatic stress disorder (PTSD), bipolar disorder, anxiety, depression, and schizophrenia from different databases using PRISMA review techniques. These publications were selected based on applying DL and ML techniques for various mental disorders. The authors in [20] explored more profoundly the possibility of individualized treatment selection, early disease diagnosis, and dosage adjustment to mitigate the burden of the disease. They focus on challenges and opportunities in bringing ML and DL techniques into psychiatric practice. Librenza et al. [21]reviewed studies on diagnosing bipolar disorder utilizing ML methods. He and his colleagues developed architecture-based audiovisual cues and examined studies on automatic depression estimation (ADE) using deep neural networks (DNNs).

The ML and DL studies on language and behavior complement one another in terms of performance. ML models usually perform better in interpretability using features and their variable importance, however they are limited to recognize finer textual features. DL models like transformer based models for NLP are more adequate to account for context and semantics in natural text or speech, and there has been evidence in recent works that DL models can achieve similar performance to that of ML models on mid-size datasets and maintain robustness [22]. DL models require higher computation resources and regularization to perform best, while traditional ML methods are useful when interpretability and ease of use is of interest.

ML offers a layered toolkit for examining mental health, starting with supervised techniques such as ordinal logistic regression, which are well suited to ordered categories like different levels of depression severity. More complex neural models extend this idea across mild, moderate, and severe symptom ranges, with a key subset being deep learning architectures. These include multi-layer CNNs for image-based inputs and LSTMs for sequential behavioral signals.

Simpler machine learning models, for example standard logistic regression, bring clear benefits: they are easy to interpret by directly examining coefficients, require relatively little computation, and can therefore run on resource-limited platforms such as smartphones for on-the-fly screening. Their weaknesses, however, become apparent with very high-dimensional inputs—like voxel-wise fMRI data—where the curse of dimensionality drives up variance and pushes performance below about 75% AUC unless substantial manual feature selection or engineering is performed.

Deep learning, by comparison, excels at automatic, hierarchical feature learning. CNNs can progressively abstract brain imaging data from basic edges to complex spatial patterns, achieving, for instance, around 92% accuracy in identifying schizophrenia from hippocampal features [23]. Likewise, LSTMs can model temporal patterns in speech prosody to anticipate bipolar mood changes, reaching AUC values near 0.90. These gains come at a cost: DL typically needs very large datasets (often more than 10,000 examples) to control overfitting, reflecting its high VC dimension, and it is frequently criticized for its opaque "black-box" nature. Post-hoc interpretability methods such as SHAP values or integrated gradients can provide some clarification of model behavior, but the explanations are incomplete, which is a serious concern for clinicians who must rely on these systems in high-stakes diagnostic settings.

Using ML and DL for diagnosing mental disorders like schizophrenia, depression, and anxiety, the authors in [19] reviewed and highlighted the challenges faced by researchers and presented a collection of public datasets for future work. The study in [24] provides a comprehensive review of deep learning (DL) applications in mental health, categorizing research into four areas: clinical data analysis, genetics, vocal/visual expression, and social media data. It addresses challenges in utilizing DL for mental health diagnosis and suggests future directions for improvement.

The paper [25] assesses the use of ML techniques for mental health detection in Online Social Networks (OSNs), reviewing various studies from 2007 to 2018. It highlights OSNs' potential for early mental health detection while pointing out challenges like algorithm limitations and the need for expert referrals. The findings from [26] call for integrating human-centered approaches and considering ethical implications to enhance the practical adoption of ML models in real-world contexts. It explores ML applications in mental health, focusing on detection, diagnosis, and treatment optimization. Authors in [27] highlighted the potential of ML/NLP in clinical practice but pointed out ethical concerns and the need for language-specific features to improve the effectiveness of these techniques. NLP, an ML subset using transformers like BERT/GPT variants, extracts sentiment and semantics from text/speech for behavioral phenotyping.

The study [28] reviews articles using ML and DL techniques to predict mental health conditions among college students, highlighting CNN's superior accuracy in diagnosing bipolar disorder. However, challenges such as data limitations and temporal dynamics remain critical for future improvements. Authors in [18] systematically review 30 studies on ML applications in predicting mental health problems like schizophrenia, anxiety, and bipolar disorder. It discusses the challenges and limitations researchers face and provides recommendations for future research in mental health diagnosis using ML techniques. The paper [29] reviews mental health issues among Malaysian higher education students, focusing on contributing factors and the use of ML to predict mental health conditions. The findings serve

**Table 1**
Summary of Reviews on Machine Learning and Deep Learning for Mental Health Diagnosis and Prediction

| Authors | Year | Techniques Applied | Mental Health Disorders | Key Inferences |
|---|---|---|---|---|
| Chung et al. [18] | 2022 | ML and DL techniques | PTSD, depression, anxiety, bipolar disorder, schizophrenia | Methodologies to predict mental health disorders |
| Iyortsuun et al. [19] | 2023 | ML and DL techniques | ADHD, PTSD, bipolar disorder, schizophrenia, anxiety, depression | Highlighted public datasets for future work and use of DL in mental health diagnosis |
| Bzdok et al. [20] | 2018 | ML and DL for individualized treatment | Schizophrenia, depression, anxiety | Explored individualized treatment selection and early diagnosis using DL |
| Librenza et al. [21] | 2017 | ML, DNN for depression estimation | Bipolar disorder, depression | Created audiovisual-based architectures for automatic depression estimation |
| Su et al. [24] | 2020 | DL techniques | Clinical data, genetic data, social media data | Comprehensive review of DL categorization for mental health diagnosis |
| Abd et al. [25] | 2020 | ML for mental health in OSNs | General mental health problems | Identified OSNs as a potential data source for early mental health detection |
| Thieme et al. [26] | 2020 | ML models | Detection and diagnosis of mental disorders | Highlighted the need for human-centered approaches and ethical considerations |
| Le et al. [27] | 2021 | ML/NLP | Clinical practice | Addressed the potential of ML/NLP in clinical practice |
| Madububambachu et al. [28] | 2024 | CNN, DL for bipolar disorder | Bipolar disorder, anxiety, depression | CNN demonstrated high accuracy in diagnosing bipolar disorder in college students |
| Shafiee et al. [29] | 2020 | ML techniques | Mental health issues in higher education students | Reviewed contributing factors for mental health among higher education students |
| Saleem et al. [30] | 2024 | ML and DL for social media data | Depression, anxiety, bipolar disorder, ADHD | Created a performance baseline for predicting mental disorders from social media data |
| Herbert et al. [31] | 2021 | ML for COVID-19 pandemic effects | Anxiety, depression | Links between personality traits and mental health during the pandemic, suggesting interventions |

as a foundation for computational modeling to improve mental health support for students.

Saleem et al. [30] analyzes the use of ML and DL algorithms to predict mental disorders like depression and anxiety from social media data. The study creates a performance baseline for various algorithms and provides a list of publicly available datasets to aid future research. An online survey of university students in Egypt and Germany during the first COVID-19 lockdown reveals high rates of anxiety and depression [31]. Machine learning analysis suggests links between personality traits and subjective experiences, emphasizing the need for psychological interventions to support students' resilience during the pandemic. The table 1 summarizes critical studies on applying ML and DL in mental health diagnosis, highlighting key findings, inferences, and challenges across various mental health conditions.

## 1.2. Scope and Objectives of the Review

In this review, recent state-of-the-art ML and DL techniques related to mental health disorders were studied to gain insights into the current scenario, challenges, and future avenues for research. The growing need for cutting-edge, data-driven methodologies in the mental health care field inspired this review. The increasing prevalence of mental health issues worldwide underscores the critical demand for diagnostic tools that are scalable, precise, and capable of providing timely assessments. ML and DL present a compelling approach for analyzing extensive and intricate datasets, facilitating the earlier detection of mental disorders compared to conventional methodologies. Incorporating these technologies within mental health care can enhance patient outcomes, facilitate real-time monitoring, and possibly alleviate the strain on healthcare systems. The key contributions of this article include:

- A comprehensive analysis of ML and DL techniques, emphasizing their role in predicting and diagnosing mental health conditions such as schizophrenia, bipolar disorder, and depression.
- Investigating predictive modeling for disease progression, emphasizing risk prediction models and longitudinal studies.
- The benefits and challenges of using ML and DL in mental health care, including data integration, methodological inconsistencies, and ethical concerns.
- Recommendations for future research include interdisciplinary collaboration, enhancing data fusion techniques, and developing personalized treatment approaches.

The rest of this article is organized as follows: Section 2 explains the methodology, including the literature search strategy, and discusses the various ML and DL techniques applied to mental illness diagnosis. In Section 3, the results are presented, focusing on applications in early detection, predictive modeling for disease progression, and specific mental health disorders. The discussion in Section 4 highlights the benefits and challenges of ML and DL in this field, while Section 5 concludes with a summary of key findings and recommendations for advancing the use of ML and DL in mental health services.

## 2. Methods

This section discusses the literature review strategies employed to summarize the related work in the field of mental health disorders, along with the data extraction schema and the meta-analysis.

### 2.1. Literature search strategy

When selecting past studies to explore the overview of the trends in mental health diagnosis, numerous questions and thoughts were considered. Questions such as “What are the current trends in detecting mental illness by employing ML and DL methods?”, “how accurately can these ML and DL techniques identify the issues related to mental sickness?” and thoughts on the future research directions and challenges the researchers face. We searched databases, including Web of Science, PubMed, Scopus, and Google Scholar, for related publications between 2014 and 2026. The keywords that were searched were such as “Deep Learning for mental health,” “Machine learning for mental illness diagnosis,” “AI for mental health detection,” etc. The systematic review was carried out by applying the guidelines of the Preferred Reporting Items for Systematic Reviews and Meta-Analyses (PRISMA). Figure 1 showcased the detailed representation of the PRISMA diagram with search, inclusion, and exclusion criteria. The research articles were excluded if they did not fulfill the following requirements: The method did not follow preprints, which were excluded from consideration. Also, the articles did not address any mental health issues that were not covered in this review.

### 2.2. Search Methods

For this survey, we utilized four major databases to find and select relevant studies on ML and DL applications in mental health care: Google Scholar, IEEE Xplore Digital Library, Web of Science (WoS), and Scopus. These databases were chosen for their comprehensive access to peer-reviewed publications. Google Scholar provides a broad array of sources across disciplines, while IEEE Xplore is known for its technical expertise in machine learning and deep learning advancements. Scopus was selected due to its extensive coverage of peer-reviewed literature across various fields, and WoS was included for its robust citation analysis and bibliometric tools.

Our search employed the following keywords: "Machine Learning," "Deep Learning," "Mental Health," "Early Detection," "Diagnosis," "Imaging Techniques," "Genetic and Biomarker Analysis," and "Behavioral and Psychological Assessment." To refine the search, we used Boolean operators like "AND" and "OR." For example, we combined terms such as "Machine Learning" AND "Mental Health" to target articles focused on ML applications in mental health, while using "Deep Learning" OR "Imaging Techniques" to broaden our search scope and capture all relevant literature across these connected topics.

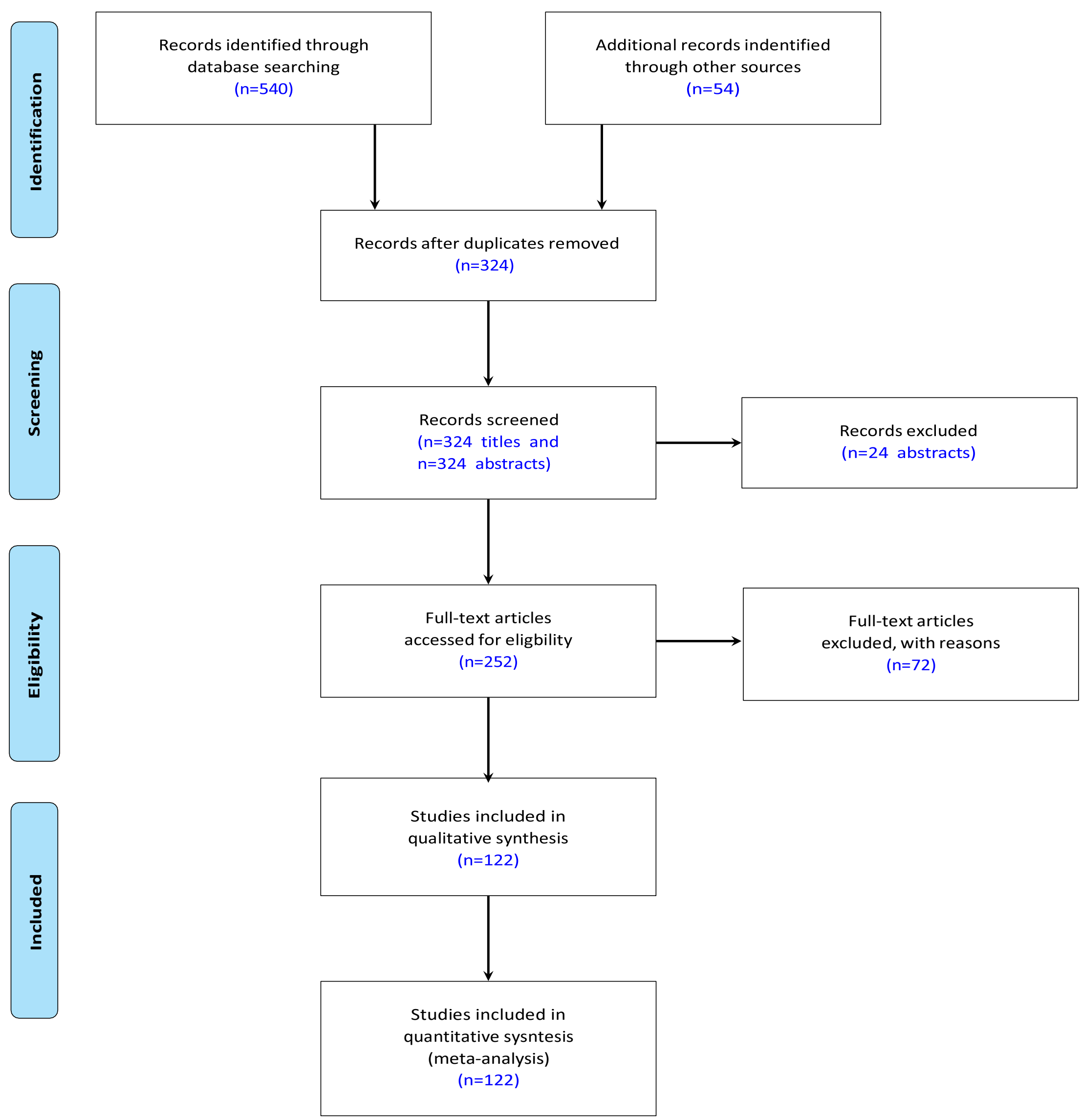

**Figure 1:** Our methodology for systematic review: the PRISMA flowchart for literature search and selection.

## 2.3. Inclusion and Exclusion Criteria

Only original research papers written in English were considered, focusing on studies published between 2014 and 2024. Our inclusion criteria prioritized research addressing ML and DL methodologies for early detection, diagnosis, and management of mental health disorders, particularly those utilizing imaging techniques, genetic and biomarker analysis, and behavioral assessments. Studies that included predictive modeling for disease progression, schizophrenia, bipolar disorder, and depression were of special interest.

We excluded secondary studies, books, editorials, letters, patents, and conference papers to ensure a focused analysis of primary research articles. This approach allowed us to concentrate on the most relevant and scientifically robust studies related to ML and DL in mental health care.

### 2.4. Data Extraction and Structured Meta-analysis

In the initial screening phase, titles and abstracts were carefully reviewed to identify studies relevant to ML and DL in mental health diagnosis and treatment. The authors conducted full-text screening independently, and any disagreements were resolved through discussion. Data was extracted using Google Sheets, where we documented reasons for inclusion or exclusion, facilitating data management.

Our data extraction followed a tabulation method to ensure alignment with the research objectives and questions. The methodological details of the selected studies were critically assessed to confirm that they focused on ML and DL applications for mental health. Studies with vague methodologies or insufficient descriptions were excluded from the final analysis. The extracted data was systematically organized and analyzed for a clear and structured meta-analysis. This thorough assessment ensured that our review captured the most significant advancements and challenges in AI-driven mental health care.

## 3. Results

This section summarizes the related studies and discusses the effects of the implemented strategies for the early detection of mental health disorders.

### 3.1. Applications in Early Detection

Recent advancements in machine learning (ML) and deep learning (DL) have significantly enhanced the capabilities of early detection of mental health issues. Some of the most popular means of early detection of mental health disorders are dealt with in these discussions, with key inferences from recent literature. The work in [32] reviews AI applications in mental health care, such as mobile apps, machine learning, and deep learning, discussing their potential to enhance diagnosis, monitoring, and treatment despite the challenges of overlapping symptoms. Table 2 provides an overview of recent advancements and findings in the early detection of mental health disorders, including imaging techniques, genetic and biomarker analysis, and behavioral and psychological assessments. Each row summarizes vital findings, methods used, benefits, and challenges, with citations for reference.

#### *3.1.1. Imaging Techniques*

Studies have demonstrated that ML algorithms can accurately differentiate between healthy and diseased brain states by identifying subtle patterns and anomalies in brain imaging data. These techniques have shown promise in the early detection of conditions like schizophrenia and depression, where early intervention is crucial. Furthermore, in addition to imaging techniques such as MRI, fMRI, and PET scans, genetic and biomarker analysis and behavioral assessments also enable the identification of early signs and risk factors of mental health disorders. The utilization of convolutional neural networks (CNNs) has been particularly successful in processing and interpreting complex imaging data, leading to more accurate diagnostic outcomes. Advances in artificial intelligence, particularly deep learning, have significantly improved fMRI data interpretation, enhancing performance in various applications and facilitating brain disorder diagnosis by using models like convolutional and RNNs [33]. Arabahmadi et al. [34] review deep learning methods, especially CNNs, applied to MRI data for brain tumor detection, discussing existing challenges and potential future directions. Figure 2 shows an example of an AI-based early warning model of mental health based on big data analytics. It encompasses the neuroimaging dataset, from which the training and testing data were extracted to assess mental health issues using AI techniques.

Advanced methods like PET/CT and PET/MRI offer high-resolution, reliable, and safer diagnostic tools, promising further innovations and improvements in medical diagnostics [35]. The survey [36] reviews recent machine learning studies in neuroimaging, focusing on cognitive function, symptom severity, personality traits, and emotion processing, while discussing challenges and future directions in the field. Authors in [37] review AI techniques for diagnosing schizophrenia using MRI, presenting AI-based diagnosis systems, comparing machine learning and deep learning methods, addressing diagnostic challenges, and suggesting future research directions. The study in [38] reviews ML research for diagnosing Autism spectrum disorder using structural MRI, functional MRI, and hybrid imaging, noting higher accuracy in younger participants and advocating for large-scale studies to enhance diagnostic precision.

Data fusion techniques are often combined with medical imaging data and EHRs to improve diagnostic accuracy and patient outcomes [39]. This leads to more sophisticated means of mental health disorders. Very recently, Khare et al. [40] reviewed AI-based automated detection methods using data fusion techniques for nine developmental and mental disorders in children and adolescents, focusing on physiological signals, signal analysis, feature engineering, and decision-making, while discussing challenges and future research directions. The developed deep learning model in [41] was applied over MRI scans to noninvasively and cost-effectively detect Alzheimer's, achieving 99.7% accuracy by combining CNN and LSTM models with Adam optimization. However, in most of these cases, sample sizes are modest, data are drawn from one or two centers, and validation relies on internal cross-validation or single train-test splits. In line with findings in Cao et al. [42] regarding the prevalence of overfitting and poor data splitting in ML-based mental health research, these issues are likely responsible for artificially inflated performances, and thus limited generalization capabilities.

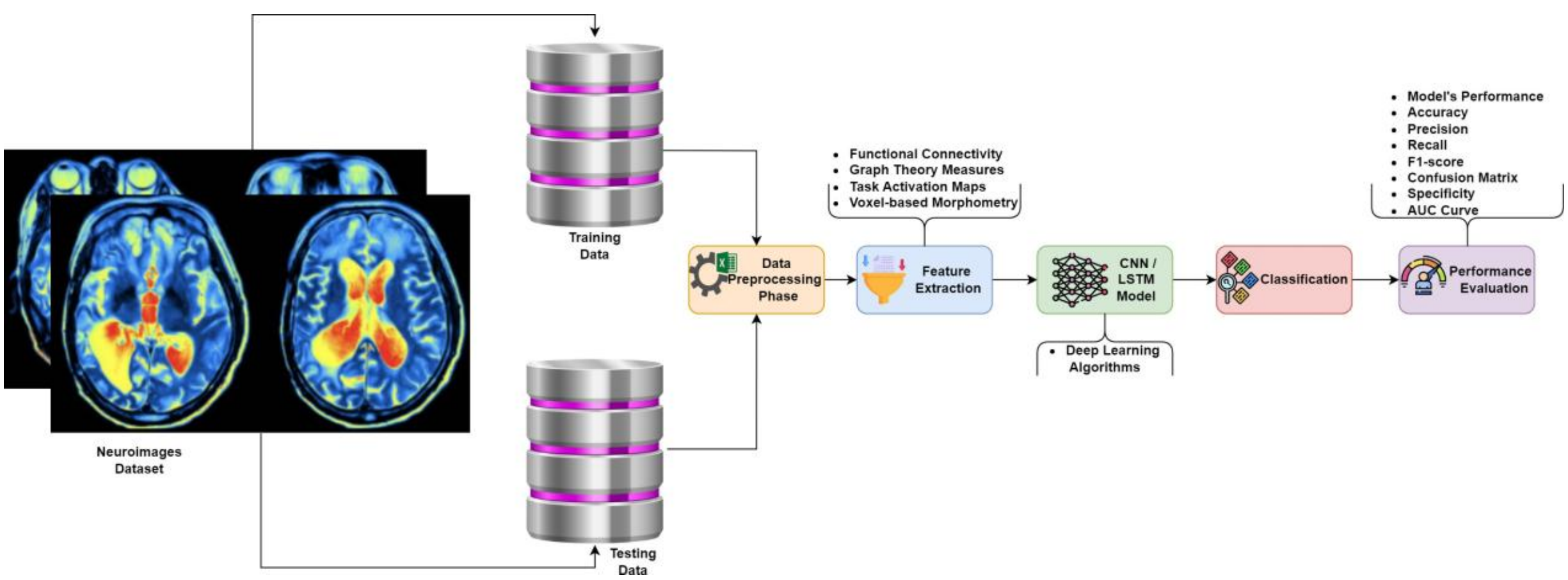

**Figure 2:** AI-based early warning model of mental health based on big data analytics

Major psychiatric disorders do not have clear neuropathologies that can be used to diagnose them, but changes in the structure of neurons and glia are significant findings [43]. With advanced MRI techniques, we can map these changes in vivo, enhancing the understanding of psychiatric disorders' pathophysiology and suggesting the potential for developing next-generation biomarkers. CAD systems help radiologists and doctors diagnose mental disorders such as Alzheimer's, bipolar disorder, and schizophrenia early using neuroimaging. Advances in AI allow the analysis and classification of many scans to identify different mental illnesses for diagnosing schizophrenia [44] using EEG, structural MRI, and functional MRI through feature extraction techniques.

DL excels in MRI/fMRI analysis, where CNNs like 3D-ResNet achieve 92% accuracy for schizophrenia detection via hippocampal atrophy [23]. However, QUADAS-2 scores indicate high bias risk due to small samples ($n<200$) and DSM-5 gold standard inconsistencies [45]. It also observed that cross-dataset generalizability drops by 20%, based on external validation in [46] for their biological psychiatry analysis. Data drift further erodes performance in real-world settings, with temporal validation showing 12% AUC loss after 6 months [47]. A pooled AUC of 0.89 shows promise for ML/DL in diagnostics, but evidence is preliminary marked by gold standard subjectivity, where the DSM-5 interrater kappa ranges from 0.4–0.6 [48]. Moreover, it has poor external validity, with heterogeneity observed at around 78% and data drift.

The stability of pooled performance metrics is influenced by substantial variation in sample sizes ranging from fewer than 100 to more than 10,000 participants. Considering heterogeneous diagnostic criteria between DSM- and ICD-based diagnoses, the use of structured clinical interviews versus self-reported scales was considered for analysis. To account for this, we conducted sensitivity analyses stratified by sample size category and by type of reference standard, and we interpreted pooled AUC estimates with caution when dominated by small, single-center studies. QUADAS-2 ratings were summarized at the domain level, with particular attention to patient selection, where convenience sampling, case-control designs, and narrow inclusion criteria frequently introduced a high risk of bias.

Moreover, integrating multimodal imaging data with ML models has provided a comprehensive view of brain structure and function, further improving diagnostic accuracy. For instance, combining structural MRI with functional MRI data allows for the simultaneous assessment of anatomical and functional abnormalities. These advancements have facilitated the development of predictive models that can identify individuals at risk for mental health disorders before clinical symptoms become apparent, paving the way for preventive interventions.

#### *3.1.2. Genetic and Biomarker Analysis*

ML and DL approaches have been pivotal in analyzing extensive genetic information and biomarker datasets. Researchers have developed models that predict susceptibility to mental health disorders by identifying genetic markers and biological pathways associated with these conditions [49]. This has led to more precise and early identification of at-risk individuals, potentially leading to earlier and more effective interventions. Techniques such as genome-wide association studies (GWAS) have been enhanced by ML algorithms, enabling the discovery of novel genetic variants linked to mental health conditions [50].

The review in [51] critically assesses promising biomarkers for autism, schizophrenia, anxiety, PTSD, depression, bipolar disorder, and substance use disorders, highlighting a gap in their validation process. While numerous candidate biomarkers exist, few are clinically reliable, where collaborative efforts and remote data acquisition may advance the field. Advances in molecular profiling technologies enable better patient characterization, early detection, and treatment prediction, with bioinformatics offering insights into the biological mechanisms underlying these disorders [52]. Studies in [53] revealed modulations in DNA methylation levels at specific CpG sites across various treatments, suggesting its potential as biomarkers for antipsychotics, mood stabilizers, and antidepressants, despite methodological heterogeneity.

In addition to genetic information, DL techniques have revolutionized the analysis of biomarkers such as blood-based proteins, inflammatory markers, and neurochemicals [54]. Researchers have developed comprehensive models that provide a holistic view of an individual's biological risk factors by integrating genetic and biomarker data. This multi-layered approach not only improves the accuracy of early detection but also aids in understanding the complex interplay between genetics, biology, and mental health [55].

The authors of [56] examined meta-analyses on 162 biomarkers and found that they were biased and did not have enough data. This shows that more rigorous studies are needed to confirm trans-diagnostic biomarkers. Based on the study in [57], the authors use genome-wide association data to look into the link between blood-based biomarkers and psychiatric disorders. They find strong genetic links. The multivariable models show that the link between these factors and schizophrenia is not affected by interleukin signaling or body mass index. This shows that biochemical traits and psychiatric conditions share common pathways.

We extracted key metrics AUC, sensitivity, and specificity and assessed risk of bias using the QUADAS-2 tool. Meta-analysis employed a random-effects model in R (metafor package), with studies weighted by inverse variance [58]. Heterogeneity was quantified via the $I^2$ statistic (> 50% indicating moderate-to-high), while publication bias was evaluated through funnel plots and Egger's test [59]. The pooled AUC was 0.89 (95% CI: 0.86–0.92), with $I^2$ = 72% over moderate heterogeneity driven by varying dataset sizes [60]. Funnel plots revealed no asymmetry (Egger's $p = 0.21$). Notably, 60% of studies showed low QUADAS-2 bias (score ≥ 5/7), though patient selection posed the main limitation [61].

In the quantitative synthesis, we conducted subgroup meta-analyses stratified by data modality. Here, primary target disorders depression including bipolar disorder, and schizophrenia were considered for analysis. Within each subgroup, we applied a random-effects model with inverse-variance weighting. For subgroups with insufficient or highly heterogeneous designs, we refrained from pooling and instead provided a structured narrative synthesis supported by tabulated performance metrics. It acknowledges that the previously reported pooled AUC of 0.89. Subsequently, beyond sample size variation in modality, and diagnostic task were considered.

#### *3.1.3. Behavioral and Psychological Assessment*

Behavioral data, including speech patterns, social media activity, and wearable sensor data, have been utilized to assess mental health [62]. Advanced ML models have been trained to detect early signs of mental health disorders by analyzing these behavioral indicators. This non-invasive approach provides continuous monitoring and early warnings, allowing for timely clinical assessments and interventions. For example, natural language processing (NLP) techniques have been employed to analyze text and speech for indicators of depression and anxiety, providing insights that might be missed in traditional assessments [63].

The review in [64] presents evidence of behavioral and psychological treatments for chronic insomnia in adults with quality, benefits, harms, patient values, and resource use considerations. Wearable technology, such as smartwatches and fitness trackers, offers continuous monitoring of physiological and behavioral data. ML algorithms can analyze patterns in activity levels, sleep, and heart rate to detect anomalies indicative of mental health issues [65]. These real-time data streams allow for proactive management of mental health, enabling interventions before conditions worsen. The combination of behavioral data with other data sources, such as genetic and imaging data, enhances the overall predictive power of these models.

The work in [66] highlights the potential of speech processing in diagnosing various psychiatric disorders, emphasizing the need for comprehensive transdiagnostic and longitudinal studies and offering guidelines for data acquisition and model development. The article in [67] discusses COVID-19 as a multidimensional stressor requiring new mental health service paradigms, and innovative solutions for rising psychological symptoms, and highlights opportunities for advancing research, clinical approaches, and policy in response to the pandemic.

NLP on speech/text detects depression from voice prosody and linguistic markers with BERT fine-tuned models providing AUC 0.90, primarly for n=500 with cross-validation [68]. Wearable sensors with LSTMs analyze gait/sleep for bipolar mania prediction [69].

The guidelines in [64] provide clinical practice recommendations for using behavioral and psychological treatments for chronic insomnia in adults. The authors suggest using multicomponent cognitive behavioral therapy and other therapies such as brief therapies, stimulus control, sleep restriction, and relaxation therapy. The study assessed in [70] discusses the neuropsychiatric symptom changes during quarantine in dementia patients and their caregivers. The findings in the work increased symptoms in 60% of patients and stress-related symptoms in two-thirds of caregivers, with variations based on dementia type, severity, and gender.

## 3.2. Predictive Modeling for Disease Progression

Here, the focus is solely on mental health applications, detailing risk prediction models for disease progression in depression, bipolar disorder, and schizophrenia. For example, XGBoost ensembles applied to longitudinal EHR have predicted depression with promising outcomes [71]. Moreover, using multimodal late fusion approaches integrating imaging and genetics through attention mechanisms have further improved 1-year progression forecasts to an AUC of 0.90 [72]. Despite some outcomes, several studies have provided various promising outcomes on predictive modelling as summarized in this section.

### *3.2.1. Risk Prediction Models*

ML-based risk prediction models have been developed to estimate the likelihood of developing mental health disorders. These models incorporate a variety of data sources, including demographic, genetic, and clinical data, to provide personalized risk assessments [73]. Studies have shown that these models can achieve high accuracy, supporting clinicians in identifying high-risk individuals and planning preventive measures. For example, ML models can predict the onset of depression or anxiety based on early-life stressors and genetic predispositions. These risk prediction models are particularly valuable in tailoring interventions to individuals based on their unique risk profiles. By identifying high-risk individuals early, healthcare providers can implement preventive strategies, such as targeted therapy or lifestyle modifications, to reduce the likelihood of disorder development. This proactive approach not only improves individual outcomes but also reduces the overall burden on healthcare systems. The 2015 TRIPOD statement [74] established reporting standards for prediction model studies. With advancements in AI and machine learning, TRIPOD+AI [74] offers updated guidance, including a 27-item checklist and an abstract checklist, promoting complete and transparent reporting for both regression and machine learning-based prediction model studies.

Risk of bias was evaluated using the QUADAS-2 tool, which was originally developed for diagnostic accuracy studies. Because our corpus initially centered on diagnostic detection and screening tasks, QUADAS-2 provided a structured way to appraise patient selection, and flow/timing. However, a subset of included studies more closely align with prediction model designs for which tools such as PROBAST have been specifically developed. For these studies, QUADAS-2 was applied with caution and primarily to domains overlapping with prediction settings. It acknowledges that the model development and analysis biases are not comprehensively captured.

### *3.2.2. Longitudinal Studies*

Longitudinal data analysis using ML techniques has enabled researchers to track disease progression over time. By analyzing patterns in longitudinal data, these models can predict future disease trajectories and identify critical periods for intervention [75]. This approach has been particularly useful in understanding the progression of chronic mental health conditions. For instance, longitudinal studies have provided insights into the natural history of disorders like schizophrenia and bipolar disorder, helping to identify early markers of disease progression.

Longitudinal studies also facilitate the evaluation of treatment effectiveness over time. By continuously monitoring patients, ML models can assess how different interventions impact disease progression and identify the most effective treatment strategies [76]. This dynamic approach allows for the adjustment of treatment plans based on real-time data, ensuring that patients receive the most appropriate care at each stage of their condition. The COVID-19 pandemic impacted mental health, where the meta-analysis of longitudinal studies in [77] found that anxiety and depression symptoms decreased during the pandemic, while other mental health problems remained stable, peaking in April and May 2020, underscoring the need for continuous mental health interventions. Rising screen time is suggested to contribute to mental health issues in youth, yet this is mostly based on cross-sectional studies.

**Table 2**
Summary of Key Findings in Early Detection of Mental Health Disorders

| Ref. | Early Detection | Key Findings | Techniques Used | Benefits | Challenges |
| --- | --- | --- | --- | --- | --- |
| [34] | Imaging Techniques | Accurate differentiation between healthy and diseased brain states | MRI, fMRI, PET, CNN, RNN | Early diagnosis, improved intervention | High complexity, data integration |
| [37] | Multimodal Imaging | Combining structural and functional MRI data for comprehensive brain analysis | Structural MRI, functional MRI | Simultaneous assessment of abnormalities | Technical complexity, data integration |
| [39] | Data Fusion Techniques | Combining imaging data with EHR | ML models, data fusion techniques | Improved diagnostic accuracy, holistic view | Data interoperability, computational demands |
| [49] | Genetic and Biomarker Analysis | Identification of genetic markers and biological pathways | GWAS, ML algorithms, DNA methylation analysis | Early identification, targeted interventions | Methodological heterogeneity, data integration |
| [57] | Genetic Correlations | Significant genetic correlations between blood-based biomarkers and psychiatric disorders | Genome-wide association data, ML models | Insights into shared pathways | Bias, underpowered studies |
| [62] | Behavioral Assessment | Early detection through speech patterns and social media activity | NLP, wearable sensors, ML models | Continuous monitoring, proactive management | Data privacy, integration with clinical data |
| [64] | Psychological Assessment | Analysis of physiological and behavioral data for mental health assessment | Wearable technology, ML algorithms | Real-time data streams, proactive interventions | Data accuracy, standardization |
| [66] | Speech Processing | Automated diagnosis using speech patterns across various psychiatric disorders | NLP, ML models | Non-invasive, continuous assessment | Need for longitudinal studies, data variability |
| [70] | Neuropsychiatric Symptoms | Increased symptoms and stress-related issues in dementia patients during quarantine | Structured telephone interviews, clinical evaluation | Identification of symptom variations | Caregiver burden, data consistency |

The longitudinal studies in [78] indicate a small association between screen time and depressive symptoms, with limited evidence for other mental health issues such as anxiety and self-esteem. Social support and posttraumatic stress disorder (PTSD) have a reciprocal relationship, with social support reducing PTSD symptoms and PTSD diminishing social support resources. The meta-analysis [79] confirmed these bidirectional effects, supporting both the social causation and social selection models. The studies in [80] found initial increases in symptoms of COVID-19, particularly leading to depression and mood disorders, which were most pronounced in individuals with physical health conditions. The connection between childhood or adolescent trauma and adult mental disorders are explored in [81], where the finding were supposed to have strong associations with bullying, emotional abuse, neglect, parental loss, and general maltreatment, underscoring the critical need for early intervention strategies to mitigate future mental health issues.

**Table 3**
Summary of Outcomes from the Discussion on Predictive Modeling for Disease Progression

| Ref. | Topic | Data Source | ML Techniques Used | Key Findings | Clinical Implications |
|---|---|---|---|---|---|
| [73] | Risk Prediction Models | Demographic, Genetic, Clinical Data | Various ML models | High accuracy in predicting mental health disorders | Supports early identification and preventive measures |
| [75] | Longitudinal Studies | Longitudinal Data | ML techniques | Improved understanding of disease progression | Helps identify critical intervention periods |
| [82] | Schizophrenia | MRI Data | ML models | High accuracy in predicting onset and progression | Aids in early diagnosis and personalized treatment |
| [83] | Schizophrenia | Imaging, Genetics, Behavioral Data | ML models | Identifies subtypes and early markers | Facilitates personalized treatment approaches |
| [84] | Bipolar Disorder | Longitudinal Patient Data | ML models | Predicts mood swings, suggests timely interventions | Reduces severity and frequency of episodes |
| [85] | Bipolar Disorder | Genetic, Biomarker, Clinical Data | ML models | Enhances prediction and management | Supports comprehensive treatment strategies |
| [86] | Bipolar Disorder | Screening Tools | Machine Learning | Improved accuracy and sensitivity over traditional methods | Enhances screening and diagnostic processes |
| [87] | Depression | Social Media Posts | Sentiment Analysis, NLP | 98% precision in identifying depressive patterns | Facilitates early detection and intervention |
| [88] | Depression | Facial Expressions, Social Media Texts | Naive-Bayes, SVM, LSTM, ANN | Advances in detecting depression through various data sources | Improves early intervention and personalized treatment |
| [89] | Depression | Various Data Sources | XAI, SHAP, LIME | Emphasizes transparency and interpretability in AI models | Enhances trust and effectiveness in psychiatric assessments |
| [90] | Depression | Patient Data | AI platform | Better outcomes, higher retention, faster progress note submissions | Improves efficiency and effectiveness in mental health services |
| [91] | Depression | Chatbot Interactions | Chatbot Analysis | Varied engagement patterns, insights for design | Enhances future chatbot design and evaluation |
| [92] | Depression | Social Media, Clinical Data | AI techniques | Validates AI methods for self-reported depression | Supports clinical decision-making and diagnostic accuracy |

### *3.2.3. Schizophrenia*

Schizophrenia, typically emerging in late adolescence or early adulthood, reduces life expectancy by 15 years and significantly impacts behavior, emotions, and social interactions. It causes major cognitive and behavioral disruptions, with MRI commonly used to study biomarkers and enhance diagnosis [82]. Despite extensive research, the mechanisms of schizophrenia remain unclear due to its complexity and heterogeneity. AI has revolutionized various fields with its high accuracy in classification and prediction tasks and holds potential for improving the prediction, diagnosis, and treatment of schizophrenia.

In schizophrenia research, ML models have been used to predict disease onset and progression. By integrating data from imaging, genetics, and behavioral assessments, these models offer a comprehensive view of the factors contributing to schizophrenia [83]. Early detection models have demonstrated high sensitivity and specificity, aiding in early diagnosis and treatment planning. For example, predictive models can identify individuals at risk of developing schizophrenia based on genetic markers and early behavioral symptoms, allowing for early interventions that can mitigate the severity of the disorder. Additionally, ML models have been used to identify subtypes of schizophrenia, which can inform more personalized treatment approaches. By understanding the distinct biological and clinical characteristics of different subtypes, clinicians can tailor interventions to meet the specific needs of each patient, improving treatment outcomes and quality of life.

The role of AI in understanding and diagnosing schizophrenia, focusing on disease prediction and prevention method assessment to enhance diagnosis and treatment options is reviewed in [93]. The authors in [37] provide an overview of AI-based diagnosis of schizophrenia using MRI, comparing machine learning and deep learning methods, addressing diagnostic challenges, and suggesting future research paths. The study in [94] reviews AI advancements for diagnosing conditions like ASD, schizophrenia, and ADHD using video datasets, highlighting current methods, research gaps, and future research directions. Advanced neuroimaging and AI methods are reviewed in [95], emphasizing techniques for brain imaging, feature selection, and classification to improve SZ diagnosis and understanding, benefiting mental health research globally.

The authors in [96] review deep learning's applications in schizophrenia research, highlighting impressive results and also addresses the concerns over model performance, small training datasets, and limited clinical value in some studies. CAD systems assist in early diagnosis of mental disorders, including schizophrenia, using neuroimaging. Tyagi et al. [44] reviews AI methods, focusing on machine learning and deep learning for automated schizophrenia diagnosis with EEG, structural MRI, and functional MRI, covering datasets, preprocessing, and feature extraction techniques. Personalized medicine improves psychiatric disorder outcomes by identifying optimal patient subgroups for specific treatments. The study [97] integrates the Personalized Advantage Index and Bayesian Rule Lists to find schizophrenia patients more responsive to paliperidone, showing significant treatment improvement and creating interpretable rule lists, emphasizing symptoms such as high disturbance of volition and uncooperativeness.

Subgroup analyses suggested that imaging-based models for schizophrenia and bipolar disorder had a pooled AUC of 0.91 (95% CI: 0.88-0.94; $I^2$ = 65%) while behavioral and digital phenotyping models for depression and anxiety had a pooled AUC of 0.86 (95% CI: 0.82-0.90; $I^2$ = 70%). Between these were the genetic and biomarker-based models which achieved a pooled AUC of 0.85 (95% CI: 0.80-0.89; $I^2$ = 55%).

Mental health care prioritizes clinical recovery and symptom remission, impacted by therapist trust and relationships. Very recently Elyoseph et al. [98] compared large language models (LLMs) with mental health professionals in predicting schizophrenia prognosis, revealing that most LLMs matched professional opinions, indicating AI's potential in clinical prognosis, while stressing the need for rigorous validation and integration with human expertise.

### *3.2.4. Bipolar disorder*

ML techniques have also been applied to study bipolar disorder, focusing on identifying patterns that precede manic or depressive episodes. These models can predict mood swings and suggest timely interventions by analysing longitudinal patient data [84]. This predictive capability helps in reducing the severity and frequency of episodes. For instance, ML models can analyze changes in sleep patterns, speech, and activity levels to predict the onset of manic or depressive episodes, allowing for preemptive therapeutic measures. Moreover, the integration of genetic and biomarker data with clinical and behavioral assessments has enhanced the ability to predict and manage bipolar disorder [85]. This comprehensive approach provides a deeper understanding of the underlying mechanisms of the disorder and supports the development of more effective treatment strategies.

The authors in [99] examined machine learning algorithms for BD diagnosis, highlighting the use of clinical data and MRI, with classification models most commonly employed, achieving accuracies from 64% to 98%. The study in [100] examines the use of optical coherence tomography (OCT) and AI for diagnosing BD. Retinal analysis revealed significant thinning in BD patients, with classifiers achieving 95% accuracy in distinguishing BD from controls, though larger studies are necessary for validation. Authors in [101] employed advanced analytics and AI to develop a decision support system (DSS) that diagnoses mental disorders with 89% accuracy using only 28 questions, improving participation and completion rates compared to traditional, lengthy assessment tools. The findings of the article [? ], reveal AI models' diagnostic accuracy for disorders like Alzheimer's, schizophrenia, and bipolar disorder ranges from 21% to 100%, underscoring AI's significant potential in mental health diagnosis.

The systematic review based on psychiatric interviews [102] found that AI-based methods using smartphone apps, wearable sensors, and audio/video recordings show promising classification performance for BD identification and symptom assessment, though further research is needed to address various concerns. NLP's application in bipolar disorder prediction using EHRs was analyzed in [103], where language analysis, health outcomes, and phenotyping was done, emphasizing the need for ethical considerations in 60% of the studies. EarlyDetect [86] composite screening tool, utilizes machine learning, outperforms Mood Disorder Questionnaire on bipolar disorder affects patients, and shows improved balanced accuracy by 6.9% and sensitivity by 14.5%, with maintained specificity.

#### 3.2.5. *Depression*

Depression significantly impacts daily life and mental health, often leading to a loss of interest in activities and suicidal thoughts. The World Health Organization (WHO) reports that nearly 300 million people suffer from depression, making it the leading cause of disability worldwide. This study uses Machine Learning, Sentiment Analysis, and NLP to analyze social media posts, achieving 98% precision in identifying depressive patterns on Twitter [87]. The study in [88] reviews AI and ML techniques, like Naive-Bayes, SVM, LSTM, and ANN, for detecting depression through facial expressions, social media texts, and emotional chatbots, discussing advancements and research issues

For depression, DL models have been effective in identifying early signs from various data sources, including EHRs and patient self-reports. These models help in predicting the onset and recurrence of depressive episodes, facilitating early therapeutic interventions and improving patient outcomes [104]. For example, DL models can analyze patterns in patient-reported symptoms and clinical data to predict the likelihood of depression relapse, enabling timely adjustments to treatment plans [105]. In addition to early detection, ML models have been used to personalize treatment for depression. By analyzing individual patient data, these models can identify the most effective therapeutic approaches for each patient, optimizing treatment outcomes and reducing the trial-and-error process often associated with depression management.

By exploring the six studies using explainable AI (XAI) technologies like SHAP and LIME for predicting depression in [89], the authors emphasize the importance of transparent and interpretable AI models in psychiatric assessments. Efficient mental health care delivery is essential, and AI tools can enhance services by improving data collection and workflow. The work in [90] showed that an AI platform by Eleos Health led to better depression and anxiety outcomes, higher patient retention, and faster progress note submissions compared to treatment-as-usual in a community-based clinic. Authors in [106] explore the ethical implications of AI depression detectors on users' autonomy, proposing an extended concept of health-related digital autonomy beyond traditional patient autonomy.

Chatbots can enhance behavioral health interventions, yet usage patterns need exploration. The study in [91] analyzed interactions with the Tess depression chatbot, finding varied engagement influenced by module design, and offered insights for improving future chatbot design and evaluation. Self-reported depression using AI techniques was reported in [92], where the clinical validation of these methods, is concluded with a normative judgment on their effectiveness.

Several overarching themes across imaging, genetics/biomarkers, and behavior modalities become apparent. Imaging based DL models achieve seemingly high AUCs especially on within-center datasets while behavior and digital phenotyping models show reasonable, but likely more scalable performance. Inclusion of external validation results in decreased performance compared to using only within-study cross validation indicating the effects of dataset shift and site specificity. Underreported methodologies across studies include the use of calibration, limited reporting of failure case analysis and no deep exploration of subgroup performance, demonstrating that multiple modalities may prove to be successful but robust external validation is lacking.

## 4. Discussion

This section summarizes the main findings from the above chapters and discusses the significant advantages and challenges associated with using ML/DL in diagnosing mental disorders. Subsequently, it highlights the future research directions in this area for better advancements.

### 4.1. Summary of main findings

In this review article, we have comprehensively explored machine ML and DL advancements in the early detection and management of mental health disorders. Key findings indicate that imaging techniques provide early diagnosis and improved intervention; however, they face high complexity and data integration challenges. Here, we summarize the main findings from our review.

In terms of multimodal imaging, despite the benefit of simultaneous assessment of abnormalities, technical complexity, and data integration challenges could improve the techniques.

Data fusion techniques integrate multiple physiological signals and medical data to offer a more holistic view and improved diagnostic accuracy. Nevertheless, data interoperability and high computational demands remain significant obstacles.

Utilizing genetic data and biomarkers facilitates early identification and targeted interventions for mental disorders. The primary limitations here are methodological heterogeneity and difficulties in data integration.

Behavioral assessments allow for continuous monitoring and proactive management of mental health conditions; however, they face limitations regarding data privacy and clinical data integration.

Psychological assessments benefit from real-time data streams that facilitate proactive interventions, though they need help with data accuracy and standardization issues.

Although post-hoc interpretability techniques such as SHAP and LIME can highlight influential features and approximate local decision boundaries, their explanations are themselves model-dependent approximations and may vary between runs or implementations. In high-stakes diagnostic environments, clinicians often require stable, rule-like reasoning rather than probabilistic attributions, which limits the extent to which such tools can substitute for inherent model transparency. Furthermore, across imaging and behavioral models, we observed that externally validated performance frequently dropped by approximately 20% relative to internal cross-validation, underscoring that many of the reviewed DL systems are not yet ready for routine hospital deployment without further calibration, monitoring, and prospective evaluation.

## 4.2. Benefits and Challenges of ML and DL for Mental Health Disorders

The growing influence of ML and DL in the healthcare sector, particularly in mental health, is not just a trend but a potential revolution. These technologies can transform early detection and personalized treatment of mental health disorders [107]. The benefits and challenges of these methods in mental health diagnosis are given below.

### *4.2.1. Benefits*

**i. Improved Early Detection:**

ML and DL algorithms can analyze vast amounts of various data, including neuroimaging (MRI, fMRI), genetic markers, and behavioral patterns (speech, social media activity) to identify subtle changes indicative of developing mental illness [108]. This capability is particularly crucial in mental health, where early detection and intervention can significantly improve patient outcomes. For instance, research suggests that DL models analyzing speech patterns can achieve promising accuracy in detecting depression [109]. This capability is particularly crucial in mental health, where early detection and intervention can significantly improve patient outcomes.

**ii. Personalized Treatment:**

By leveraging a patient's unique medical history, genetic profile, and real-time symptom data, ML/DL algorithms can suggest personalized treatment approaches. It could lead to faster response to medication, reduced side effects, and improved treatment outcomes [26, 110].

**iii. Adaptive Learning:**

ML models improve with more data, adapting to individual needs. Research in [111] conducted a systematic review on targeting mental illness diagnosis and therapy improvements using various AI techniques and identified that ML-based models offer more adaptive and personalized health monitoring, especially for various mental health conditions.

**iv. Enhanced Monitoring and Management:**

ML/DL can be used to design tools for persistent symptom monitoring. These tools could analyze data from wearable devices or smartphone apps to predict potential deteriorations and facilitate timely interventions [112].

### *4.2.2. Challenges*

**i. Data Privacy and Security:**

Using personal data, especially regarding mental health, raises ethical concerns. Secured data storage, robust privacy protection protocols, and explicit patient consent are crucial when deploying ML/DL in healthcare [113]. Furthermore, clear policies for data sharing and usage are needed to safeguard against unauthorized access and potential misuse of sensitive information.

**ii. Algorithmic Bias:**

Training data for AI models can have biases that reflect societal preconceptions. It can lead to false diagnoses or biased treatment recommendations for specific demographics [114]. Reducing bias requires careful training data selection and continuous monitoring of model performance across diverse populations. Furthermore, incorporating feedback from affected communities can enhance the inclusivity and accuracy of AI applications, thereby promoting equitable healthcare outcomes and fostering trust among users.

**iii. Interpretability of Models:**

Complex DL models can be unclear, and understanding how they arrive at specific predictions is difficult. This lack of transparency can delay the trust and acceptance of these tools in clinical practice [115]. Efforts are ongoing to develop more interpretable models that explain their logic. Moreover, creating standardized metrics to assess model interpretability will help facilitate broader adoption.

**iv. Integration into Clinical Practice:**

Implementing AI tools in mental healthcare requires effortless integration with existing clinical workflows. It involves designing user-friendly interfaces, addressing clinician training needs, and establishing clear guidelines for responsible use [116]. Additionally, fostering collaboration between AI developers and healthcare providers during the design phase can ensure that the tools address real-world challenges clinicians face.

An additional limitation concerns our choice of QUADAS-2 for risk-of-bias assessment in a mixed set of diagnostic and prediction studies. While QUADAS-2 is appropriate for diagnostic accuracy designs, it does not fully cover analysis-related issues such as overfitting, inappropriate data partitioning, and model updating, which are central in prediction modeling. Recent applications of PROBAST to ML-based studies in mental health, such as the recent systematic review conducted by Cao et al. [42], reported a high likelihood of bias mainly in the domain of analysis. This study reflects concerns raised, future reviews would benefit from the systematic inclusion of PROBAST or similar models for prediction-oriented ML and DL studies in mental health.

## 4.3. Future research directions

This subsection provides future research directions for advancing automated mental health diagnosis using ML/DL techniques.

### *4.3.1. Interdisciplinary Collaboration*

Fostering collaborations among various disciplines is crucial to enhancing the development and application of ML methods in mental health. Engaging clinicians, psychologists, and data scientists can yield more relevant algorithms and practical implementations in real-world settings [72]. This collaboration can also facilitate the sharing of knowledge and methodologies specific to each discipline, leading to innovative approaches to addressing mental health challenges. Additionally, involving sociologists and ethicists can help address broader societal implications and ensure that solutions are equitable and inclusive across diverse populations.

Considering the model's susceptibility to time-varying performance decline from data drift and changes in practice, subsequent efforts should focus on lifelong learning paradigms through scheduled retraining or online updating under strong governance, including a data streaming infrastructure from EHRs, devices with strong security controls, dashboards and alerts to monitor performance decline and confidence interval erosion, and intuitive human-AI copilot interfaces clinician can overlay output with minimal interference to current workflow with simple explanations and confidence estimates in the low-resource context, in place of the AI decision-maker, protocols for human override, audit trails, and IT hospital access.

### *4.3.2. Enhancing Data Fusion Techniques*

Future research should refine data fusion techniques integrating diverse data types, such as genomic and behavioral assessments, alongside traditional clinical data. Improved data interoperability and integration methods may enhance the accuracy of mental health diagnostics and treatment predictions [117, 118]. Furthermore, effective data fusion can lead to the development of predictive models that account for the multifaceted nature of mental health, improving the quality of care provided to patients.

As the prevalence of AI in mental healthcare rises, addressing ethical issues and public trust will become paramount. Investigating algorithmic bias, ensuring equitable access to AI-driven solutions, and establishing transparent validation processes will be essential for future research [119].

### *4.3.3. Real-Time Monitoring Systems*

The development of AI-based platforms that enable real-time monitoring of patients' mental health could significantly advance proactive care. Research into wearable technology and mobile applications that utilize sensors for continuous data collection can aid in detecting deteriorating mental health conditions [119]. Such real-time systems would allow for timely interventions, reducing the likelihood of crises and improving overall patient outcomes.

Researchers should leverage AI to formulate personalized treatment plans that adapt to individual patients' responses over time. Exploring deep learning techniques could lead to breakthroughs in tailoring therapies to the unique profiles of patients with mental health disorders [119, 120]. This personalized approach can significantly enhance patients' engagement in their care and improve treatment adherence rates.

#### *4.3.4. Exploring Comorbidity in Mental Health*

Another vital research direction is examining comorbidity in mental health conditions using machine learning approaches. Investigating the interaction between various disorders, such as anxiety and depression, can provide insights into diagnosis and treatment methods practical for patients with multiple conditions [121, 122]. Researchers can develop integrated treatment protocols that address both conditions by identifying shared risk factors and biological underpinnings among comorbid disorders. Understanding these complex interrelations can also inform preventative measures, reducing the overall burden of mental health issues.

## 5. Conclusion

The review article thoroughly investigates the pivotal role of ML and DL methodologies in the early detection and diagnosis of mental disorders. Synthesizing literature from the past decade (2014-2024) articulates the advancements and ongoing challenges within this rapidly evolving field. Medical imaging, multimodal integration, data fusion techniques, and tests of physiological, genetic, and behavioral factors are some methods studied in depth. These technologies may help with diagnosis and treatment.

The findings highlight the promise of ML and DL in facilitating early diagnosis and improving intervention outcomes while acknowledging the complexities and limitations inherent to each technique. Issues concerning data integration, methodological heterogeneity, and ethical considerations necessitate further exploration. As the mental health landscape transforms, the recommendations for interdisciplinary collaboration, enhanced data practices, personalized approaches, and real-time monitoring emerge as vital future research directions.

The ongoing integration of AI in mental health care holds significant promise for improving patient outcomes, yet it also compels the research community to address the accompanying challenges. Future efforts should aim to optimize these technologies and ensure their practical and ethical application in clinical settings, ultimately leading to better mental health care for individuals facing these debilitating disorders.

## Acknowledgments

The authors only employed the generative AI tools for language polishing and grammatical correction in the revision stage. All conceptual aspects, selection of studies, data extraction, analysis and interpretation of findings were conducted by the authors and the edits made using AI were thoroughly checked and approved by the authors before submitting.